\documentclass[a4paper,final]{IEEEtran}    % TWOCOLS

\usepackage[caption=false,font=normalsize,labelfont=sf,textfont=sf]{subfig}
\usepackage[draft]{graphicx}
\usepackage{multirow}
\usepackage{amsfonts}
\usepackage{amssymb}
\usepackage{amsbsy}
\usepackage{amsmath}
\usepackage[ansinew]{inputenc}
\usepackage{cite}
\usepackage{multirow}
\usepackage{color}
\usepackage{booktabs}
\usepackage{lscape}
\usepackage{rotating}
\usepackage{color, xcolor}
\usepackage{algorithmicx}
\usepackage{algpseudocode}
\usepackage{algorithm}
\usepackage[colorlinks,linkcolor=red,anchorcolor=blue,citecolor=green]{hyperref}
\usepackage{array}
\newcolumntype{L}[1]{>{\raggedright\let\newline\\\arraybackslash\hspace{0pt}}b{#1}}
\newcolumntype{C}[1]{>{\centering\let\newline\\\arraybackslash\hspace{0pt}}p{#1}}
\newcolumntype{R}[1]{>{\raggedleft\let\newline\\\arraybackslash\hspace{0pt}}b{#1}}
\newcolumntype{x}[1]{>{\centering\let\newline\\\arraybackslash\hspace{0pt}}p{#1}}
\newcolumntype{H}{>{\setbox0=\hbox\bgroup}c<{\egroup}@{}}
\usepackage{colortbl}
\usepackage{bm}
\usepackage{float}
\usepackage{textcomp}
\usepackage{subfig}

\usepackage{enumerate}

\usepackage{multirow}

\usepackage{soul}%% for \st strikethrough

\usepackage{pifont}% BLS: for nice tick and arrow symbols

\usepackage{arydshln}

\newcommand{\mytitle}{The Outcome of the 2022 Landslide4Sense Competition: Advanced Landslide Detection from Multi-Source Satellite Imagery}

\begin{document}

\title{\mytitle}

\author{Omid Ghorbanzadeh, Yonghao Xu, ~\IEEEmembership{Member,~IEEE}, Hengwei Zhao\IEEEauthorrefmark{2}, ~\IEEEmembership{Student Member,~IEEE}, Junjue Wang\IEEEauthorrefmark{2}, ~\IEEEmembership{Student Member,~IEEE}, Yanfei Zhong, ~\IEEEmembership{Senior Member,~IEEE}, Dong Zhao, Qi Zang, Shuang Wang, ~\IEEEmembership{Member,~IEEE}, Fahong Zhang, Yilei Shi, ~\IEEEmembership{Member,~IEEE}, Xiao Xiang Zhu, ~\IEEEmembership{Fellow,~IEEE}, Lin Bai, Weile Li, Weihang Peng, and Pedram Ghamisi, ~\IEEEmembership{Senior Member,~IEEE}
\thanks{Manuscript received XX 2021;}%\red{December} 2018;}
\thanks{O. Ghorbanzadeh, and Y. Xu are with the Institute of Advanced Research in Artificial Intelligence (IARAI), Landstra$\beta$er Hauptstra$\beta$e 5, 
1030 Vienna, Austria (e-mail: omid.ghorbanzadeh@iarai.ac.at; yonghao.xu@iarai.ac.at).\newline
\indent H. Zhao, J. Wang, and Y. Zhong are with the State Key Laboratory of Information Engineering in Surveying, Mapping and Remote Sensing, Wuhan University, 430074, China (e-mail: whu\_zhaohw@whu.edu.cn; kingdrone@whu.edu.cn; zhongyanfei@whu.edu.cn).\newline
\indent D. Zhao, Q. Zang, and S. Wang are with the School of Artificial Intelligence,
Xidian University, Xian, 710071, China (e-mail: shwang@mail.xidian.edu.cn). \newline
\indent F. Zhang is with the Data Science in Earth Observation, Technical
University of Munich, 80333 Munich, Germany.
Y. Shi is with the Chair of Remote Sensing Technology, Technical University of Munich, 80333 Munich, Germany.
X. Zhu is with the Remote Sensing Technology Institute (IMF), German Aerospace Center (DLR), 82234 Wessling, Germany and Signal Processing in Earth Observation, Technical University of Munich (TUM), 80333 Munich, Germany (e-mail: xiaoxiang.zhu@dlr.de).\newline
\indent L. Bai, W. Li, and W. Peng are with the State Key Laboratory of Geohazard Prevention and Geoenvironment Protection, Chengdu University of Technology, No.1 East Third Road, Erxianqiao, Chenghua District, Chengdu, 610059, China (e-mail: bailin@cdut.edu.cn; liweile08@cdut.edu.cn; pangdarren@outlook.com).\newline
\indent P. Ghamisi is with (1) Institute of Advanced Research in Artificial Intelligence (IARAI), Landstra$\beta$er Hauptstra$\beta$e 5, 
1030 Vienna, Austria; (2) Helmholtz-Zentrum Dresden-Rossendorf, Helmholtz Institute Freiberg for Resource Technology, Machine Learning Group, Chemnitzer Str. 40, 
09599 Freiberg, Germany (e-mail: p.ghamisi@gmail.com).
%\indent \red{XXX acknowledges YYY (fundings)}.\newline

\IEEEauthorblockA{\IEEEauthorrefmark{2}These authors contributed equally.}
}}

\markboth{IEEE Journal of Selected Topics in Applied Earth Observations and Remote Sensing, Vol. XX, No. YY, Month ZZ 202X}{Title: \mytitle}

\maketitle
\bibliographystyle{IEEEtran}%IEEEtran, IEEEbib

\begin{abstract}
The scientific outcomes of the 2022 Landslide4Sense (L4S) competition organized by the Institute of Advanced Research in Artificial Intelligence (IARAI) are presented here. The objective of the competition is to automatically detect landslides based on large-scale multiple sources of satellite imagery collected globally. The 2022 L4S aims to foster interdisciplinary research on recent developments in deep learning (DL) models for the semantic segmentation task using satellite imagery. In the past few years, DL-based models have achieved performance that meets expectations on image interpretation, due to the development of convolutional neural networks (CNNs). The main objective of this article is to present the details and the best-performing algorithms featured in this competition. The winning solutions are elaborated with state-of-the-art models like the Swin Transformer, SegFormer, and U-Net. Advanced machine learning techniques and strategies such as hard example mining, self-training, and mix-up data augmentation are also considered.
%The first team performed a series of experiments using a multi-scale U-Decoder, the Swin Transformer, and EfficientNetV2 encoders, SegFormer with self-attention operations, soft cross-entropy loss, and pseudo labels to achieve the best F1 score of 73.07\% in the competition. 
Moreover, we describe the L4S benchmark data set in order to facilitate further comparisons, and report the results of the accuracy assessment online. The data is accessible on \textit{Future Development Leaderboard} for future evaluation at \url{https://www.iarai.ac.at/landslide4sense/challenge/}, and researchers are invited to submit more prediction results, evaluate the accuracy of their methods, compare them with those of other users, and, ideally, improve the landslide detection results reported in this article.  
\end{abstract}


\section{Introduction}

\IEEEPARstart{L}{andslides} are a frequent natural hazard observed in mountainous terrains across the globe \cite{Guzzetti2006}. There are several mechanisms by which soil, rock, and objects located on the ground or underground on an unstable hill slope can move downward and create a landslide \cite{varnes1958landslide}. Landslides mainly occur in response to natural processes like heavy rainfalls and earthquakes or human-induced activities \cite{Lima2017}. The downward movement of the most catastrophic landslides is fast. They can travel large distances and take down everything in their path, creating scars on higher slopes and accumulating to deposition in valleys \cite{tayyebi2022two}. Landslides in mountainous areas are a problem, responsible for substantial losses, including damage to buildings and infrastructure and even fatalities \cite{branke2022extending}. The current climate changes, population growth, and rapid urbanization in areas vulnerable to natural hazards have also increased the occurrence of landslides and their consequences \cite{ghorbanzadeh2022landslide4sense}. As a result, in recent years, a considerable amount of attention has been paid to gaining a better understanding of the mechanisms of these catastrophic hazards \cite{van2022physically}. The most vital information regarding these catastrophic events is the awareness of past movements and their exact locations and extensions, ideally recorded in a landslide inventory data set \cite{branke2022extending}. Such a data set is an essential requirement for extracting advanced information, developing knowledge in the field, and predicting the unstable slopes that are prone to landslides \cite{Yang2022,liu2021landslide,xu2022feature}. Prediction maps generated from such a data set can be used for potential mitigation measures for the region under the study \cite{titti2021enough}. Therefore, a more accurate and detailed landslide inventory data set is a prerequisite for a precise disaster mitigation action \cite{Ghorbanzadeh2021}.

\begin{figure*}[th]
\centering
\includegraphics[width=\linewidth]{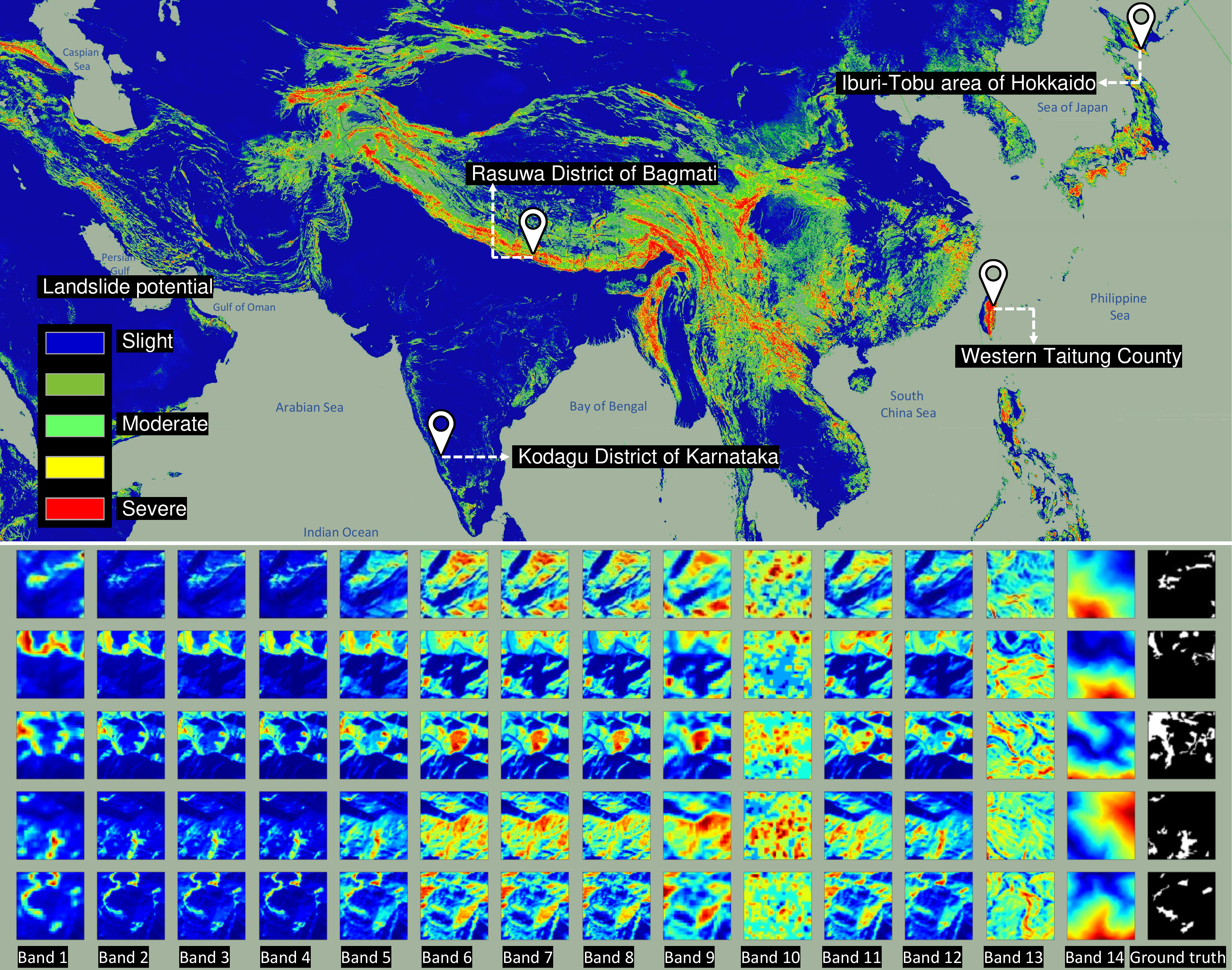}
\caption{The locations of the training sites on a global image of landslide susceptibility generated by\cite{stanley2017heuristic} and the visualization of every image layer in the $128 \times 128$ window size patches of the landslide training data set. Multi-spectral Sentinel-2 data is represented by bands 1--12, and slope and DEM data is represented by bands 13--14. The patches in the last column refer to the corresponding ground truth polygons.}
\label{fig1}
\end{figure*}

In the past decade, deep learning (DL) has gained a great deal of attention, both in computer vision and remote sensing (RS) image analyses. The application of deep learning and convolutional neural networks (CNNs) to the detection of landslides emerged in early 2019, primarily using very high resolution (VHR) \cite{ghorbanzadeh2019evaluation} and hyperspectral RS data \cite{Ye2019}. The prospect of generating landslide maps with more accuracy than can be achieved with traditional methods such as semi-automated \cite{Holbling2012} and machine learning classifiers \cite{TavakkoliPiralilou2019} has encouraged researchers in this field to develop and apply more sophisticated DL algorithms. To the best of our knowledge, no DL algorithm has been designed specifically for the distinct characteristics of landslide detection. Therefore the application of existing DL models and their variations for this task poses some new concerns, namely their transferability to new geographical areas with different landcovers and morphologies and the lack of any comprehensive open-source benchmark data set \cite{Ghorbanzadeh2021}. 

The artificial intelligence for remote sensing (AI4RS) group of the Institute of Advanced Research in Artificial Intelligence (IARAI) is a small international group of scientists working on the development and application of state-of-the-art deep learning (DL) solutions and algorithms for satellite imagery interpretation. This group has organized the Landslide4Sense (L4S) competition to foster ideas and progress in DL algorithms for the specific Earth observation application of landslide detection. The competition provides participants with a landslide benchmark data set with globally distributed multi-source satellite imagery. The benchmark data set is prepared and introduced as an explicit norm for evaluating alternative DL approaches. The training set, which is a subset of the whole benchmark data set is released and thoroughly described by Ghorbanzadeh \emph{et al.} in \cite{ghorbanzadeh2022landslide4sense}. The study evaluates this subset of the benchmark data set using 11 different state-of-the-art DL segmentation models.

\begin{figure*}[th]
\centering
\includegraphics[width=\linewidth]{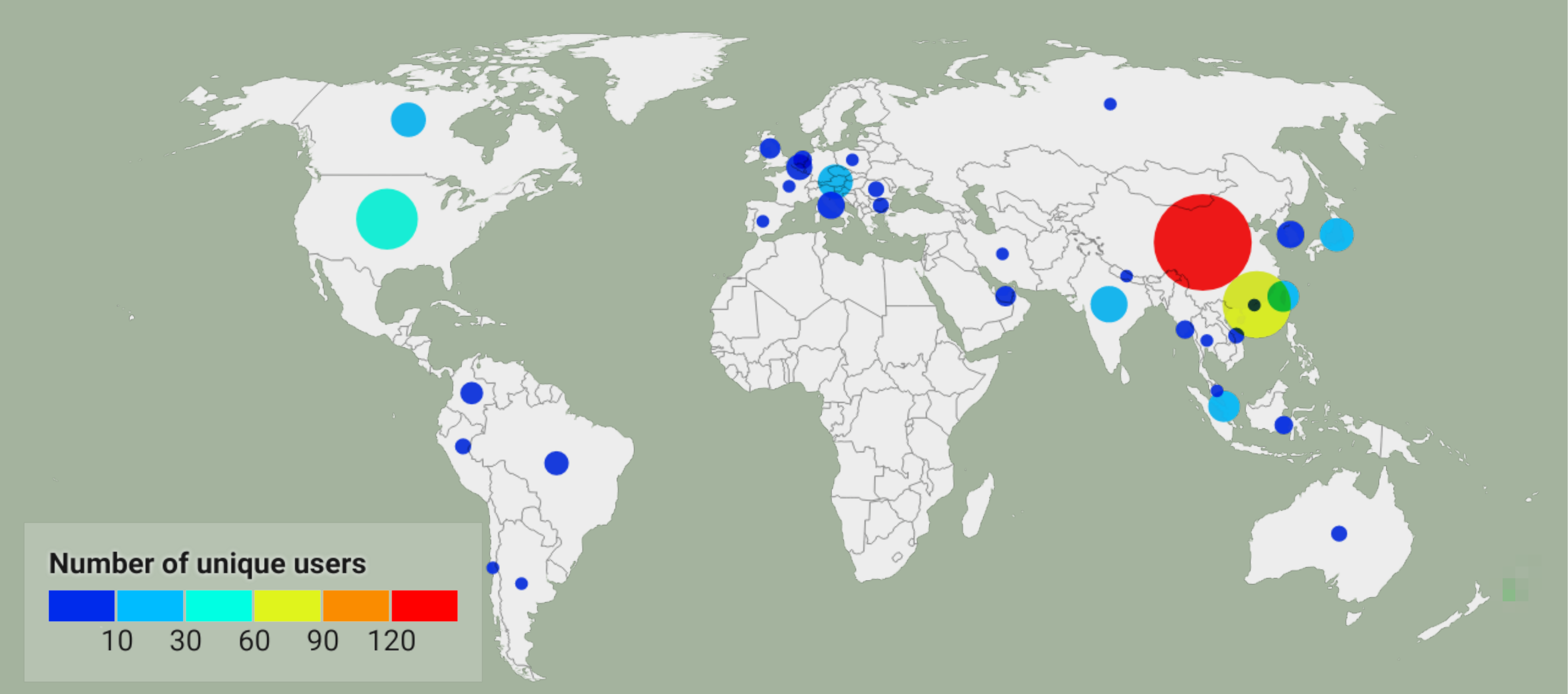}
\caption{Global distribution and number of unique users per country or region, created in \url{https://app.datawrapper.de}.}
\label{fig2}
\end{figure*}

The L4S competition fosters interdisciplinary research in computer vision, artificial intelligence (AI), and RS image analysis for image classification and landslide detection. The global objective is to build DL-based models for understanding the differentiating characteristics of landslides based on the provided optical, digital elevation model (DEM), and slope layers from freely available satellite imagery acquired by Sentinel-2 sensors and ALOS PALSAR. During the L4S competition, along with the highest accuracy assessment results, a special prize was also awarded for the most creative and innovative solution.

The competition is organized by IARAI and aims to improve automatic landslide detection DL algorithms using multi-source satellite imagery. In this competition, the main objective is the creation of landslide inventory maps using only the specified labeled landslide benchmark data set as training data.

%% outline
The main focus of this article is on the scientific outcomes of the L4S competition. The rest of the paper is organized as follows. Section~\ref{sec:data} describes the L4S benchmark data set used in the competition. Section~\ref{sec:res} provides statistics of submissions and the overall results of the competition. In the next four sections, we discuss the DL algorithms proposed by the first- to third-ranked teams and the team of the special prize. Finally, we summarize our concluding points in Section~\ref{sec:concl}.

\section{The Data and Baseline of Landslide4Sense Competition 2022}\label{sec:data}

\subsection{Data Set}
The benchmark data set for the L4S competition comprises 14 layers of data: multi-spectral data from Sentinel-2 (band1--band12), digital elevation model (DEM), and slope data from ALOS PALSAR. All 14 layers in the landslide benchmark data set are resized to the resolution of about 10m per pixel and are labeled pixel-wise to landslide and non-landslide classes. The landslide benchmark data consists of the training, validation, and test sets that encompass events occurring across a wide range of geographical locations throughout the world's mountainous regions. Specifically, only the training subset is acquired from four different sites: the Iburi-Tobu area of Hokkaido, the Kodagu district of Karnataka, the Rasuwa district of Bagmati, and western Taitung County. The data collected from these four sites provide 3,799 image patches with a size of $128 \times 128$ pixels (see Fig. \ref{fig1}). The validation and test sets contain 245 and 800 image patches of the same size, respectively, which were acquired from other geographical sites. Details about the 14 layers of the landslide benchmark data set are given below.

\begin{table}
\centering
\caption{Baseline Approach Results in the Validation and Test Phases}
\label{tab:baseline-results}
\begin{tabular}{@{}cccc@{}}
\toprule
 & Precision ($\%$) & Recall ($\%$) & F1 Score ($\%$)\\ \midrule
Validation & 51.75 & 65.50 & 57.82 \\
Test & 52.45 & 69.87 & 59.92\\ \bottomrule
\end{tabular}%
\end{table}

\begin{itemize}
\item \textbf{Sentinel-2}. The multi-spectral Sentinel-2 layers are provided in wavelengths of ultra blue, blue, green, red, visible and near infrared (VNIR), and short wave infrared (SWIR). The bands (B2, B3, B4, B8) have a spatial resolution of 10m, whereas those of (B5, B6, B7, B11, B12) and (B1, B9, B10) have a spatial resolution of 20m and 60m, respectively. This imagery is captured during cloud-free days after the event. 

\item \textbf{ALOS PALSAR}. The ALOS phased array type l-band synthetic aperture radar layers have a spatial resolution of 12.5 m and were acquired from 2006 to 2019. The Alaska satellite facility (ASF) is one of the distributed active archive centers that provides high-resolution DEM from ALOS PALSAR at no cost to the user. The slope layer is derived from the DEM ALOS PALSAR, and both DEM and slope layers are converted to 10m spatial resolution. More details about the landslide benchmark data set can be found in \cite{ghorbanzadeh2022landslide4sense}.
\end{itemize}

The task of the L4S competition is to predict landslides from the data set  provided. The labels are only provided for 3,799 image patches of the training data set. The landslide detection results are evaluated with the pixel-wise F1 Score on the landslide category in both the validation and test phases. Rankings for the competition were determined using only this accuracy assessment metric. However, competitors also received precision and recall metrics during the validation phase to get more meaningful feedback for their landslide detection results.

\subsection{Baseline}

We provided a simple baseline in our public GitHub repository prior to the start of the L4S competition.\footnote{\url{https://github.com/iarai/Landslide4Sense-2022}} A state-of-the-art DL model for semantic segmentation was implemented in PyTorch in order to provide this service. This model contains a user-configurable training script for U-Net \cite{ronneberger2015u} and the data loader for reading the training and test data sets. U-Net was first applied to biomedical image segmentation, followed by numerous semantic segmentation applications that demonstrated successful results. This model is also common for the landslide detection task and has been applied in a number of studies \cite{Ghorbanzadeh2021, Liu2020a, tang2022automatic}. U-Net comprises an encoder route capable of capturing low-level representations and a decoder route designed to capture high-level representations. As the decoder route is asymmetrical, where the vanished content of the localization is restored by using an asymmetrical design, the encoder route follows a standard CNN design assembled from consecutive convolution blocks. There is a max-pooling layer with a filter size of $2 \times 2$ and a stride of 2, after two convolutional layers with a filter size of $3 \times 3$, leveraging the rectified linear unit (ReLU) activation function in each block \cite{ronneberger2015u}. The baseline model implemented in the L4S competition includes 23 convolutional layers, of which 4 are convolutional-transpose layers. The baseline U-Net model is trained using the training data set and tested on 245 and 800 image patches of validation and test data sets, respectively. The resulting baseline accuracy for the validation and test data sets is represented in Table \ref{tab:baseline-results}. We used all 14 bands for training and testing, and no additional measurements were applied (e.g., data augmentation, pre- or post-processing. Adding any external auxiliary data such as very high-resolution images was forbidden, as specified in the L4S competition terms and conditions. The best performance of the baseline model achieves an F1 score of 59.92\% on the test set.

% \begin{table*}[]
% \centering
% \caption{Baseline results on the validation set.}
% \label{tab:baseline-results}
% \begin{tabular}{@{}cccccc@{}}
% \toprule
% \textbf{Change Class} & \textbf{NLCD difference} & \textbf{U-Net both} & \textbf{U-Net separate} & \textbf{FCN both} & \textbf{FCN separate} \\ \midrule
% Water loss & 0.1481 & 0.2751 & 0.3381 & 0.6391 & 0.6712 \\
% Tree Canopy loss & 0.1668 & 0.4828 & 0.4731 & 0.6299 & 0.6725 \\
% Low Vegetation loss & 0.2818 & 0.4769 & 0.4667 & 0.4595 & 0.5504 \\
% Impervious loss & 0.0144 & 0.2914 & 0.2669 & 0.2381 & 0.2627 \\
% Water gain & 0.0310 & 0.1577 & 0.2417 & 0.2126 & 0.1534 \\
% Tree Canopy gain & 0.0008 & 0.1478 & 0.2411 & 0.1181 & 0.1924 \\
% Low Vegetation gain & 0.1058 & 0.3510 & 0.3465 & 0.5078 & 0.5562 \\
% Impervious gain & 0.3622 & 0.5163 & 0.5142 & 0.5449 & 0.5651 \\ \midrule
% Mean IoU & 0.1389 & 0.3374 & 0.3610 & 0.4188 & 0.4530 \\ \bottomrule
% \end{tabular}%
% \end{table*} 
%

\section{Submissions and Results}\label{sec:res}

%% organization, evaluation metric

%% statistics about the contest
%% registration at Codalab

There were 439 unique users within 85 teams that submitted 7775 landslide detection results to the validation phase of the L4S competition website.\footnote{\url{https://www.iarai.ac.at/landslide4sense/challenge/}} The number of total submissions to the test leaderboard was 219 landslide detection results. We limited submissions per team to ten for the test phase. The final ranking was determined based on the highest F1 Score of each team during the test phase. Moreover, a special prize was also considered for the most creative and innovative solution in landslide detection, in the view of the L4S scientific committee. The competitors were from 37 different countries or regions. Most of the competitors were from mainland China, with 134 unique users, followed by 62 from Hong Kong, 50 from the USA, and 42 from Germany. Fig. \ref{fig2} shows the distribution and the approximate number of unique users per country or region.

The first three ranked teams that were selected based on their highest F1 Scores and the team recipient of the special prize were named winners of the L4S competition and presented their solutions during IJCAI-ECAI 2022, the 31st International Joint Conference on Artificial Intelligence, and the 25th European Conference on Artificial Intelligence at the Workshop on Complex Data Challenges in Earth Observation, CDCEO 2022. 

The four winning teams are as follows:
\begin{itemize}
    \item \textbf{1st place}: \textit{Kingdrone} team; Junjue Wang,\footnote{These authors contributed equally.\label{contri}} Hengwei Zhao,\textsuperscript{\ref{contri}} Yang Pan, Ailong Ma, Xinyu Wang, and Yanfei Zhong from Wuhan University, China.% Add the corresponding CDCEO paper later ~\cite{dfc2021msd1st}.
    \item \textbf{2nd place}: \textit{Seek} team; Dong Zhao, Qi Zang, Zining Wang, Dou Quan, and Shuang Wang from Xidian University, China.% Add the corresponding CDCEO paper later ~\cite{dfc2021msd1st}.
    \item \textbf{3rd place}: \textit{Tanmlh} team; Fahong Zhang, Zhitong Xiong, Qingsong Xu, Wei Yao, Yilei Shi, and Xiao Xiang Zhu from Technical University of Munich (TUM), Germany; German Aerospace Center (DLR), Germany.% Add the corresponding CDCEO paper later ~\cite{dfc2021msd1st}.
    \item \textbf{Special prize}: \textit{Sklgp} team; Qiang Xu, Weile Li, Lin Bai, Kai Chen, Weihang Peng, Zhenzhen Duan, and Huiyan Lu from Chengdu University of Technology, China.% Add the corresponding CDCEO paper later ~\cite{dfc2021msd1st}.
\end{itemize}

 Detailed descriptions of the four winning solutions can be found in the following sections. 
\section{First-Place Team}\label{sec:kingdrone}

\subsection{Analysis of the Characteristics of Landslide}
A progressive label refinement-based distribution adaptation landslide detection framework was proposed by the first-place team for large-scale landslide detection.
The unique characteristics of landslides create two particular challenges for large-scale landslide detection from remote sensing images: \textbf{small objects and class imbalance}, and \textbf{distribution inconsistency}.

\begin{figure}
\centering
\subfloat[Small objects]{
\label{fig:small_landslide_objects}
\includegraphics[width=0.4\linewidth]{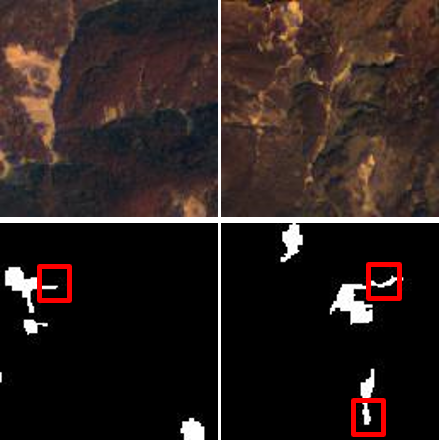}}
\subfloat[Class imbalance]{
\label{fig:class_imbalance}
\includegraphics[width=0.4\linewidth]{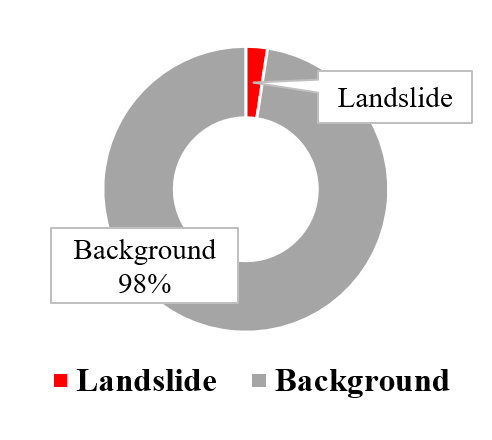}}
\caption{The problem of small objects and class imbalance.
The landslide has some smaller branches and the background has 49 times as many pixels as the landslide.}
\label{fig:first_problem}
\end{figure}

The first challenge, small objects and class imbalance, is shown in Fig.~\ref{fig:first_problem}.
In remote sensing images, the morphology of landslides is very complex, especially with many small branches, which belong to small objects (Fig.~\ref{fig:small_landslide_objects}).
Furthermore, the landslide is not the dominant ground object in large-scale remote sensing images, as shown in Fig.~\ref{fig:class_imbalance}, which illustrates the statistical result of the training data set in which the proportion of pixels occupied by the landslide is only $2\%$, and the number of pixels of other ground objects (background) is 49 times that of the landslide.
Both of these challenges, of small objects and class imbalances, lead to lower recall scores.

\begin{figure}
\centering
\includegraphics[width=\linewidth]{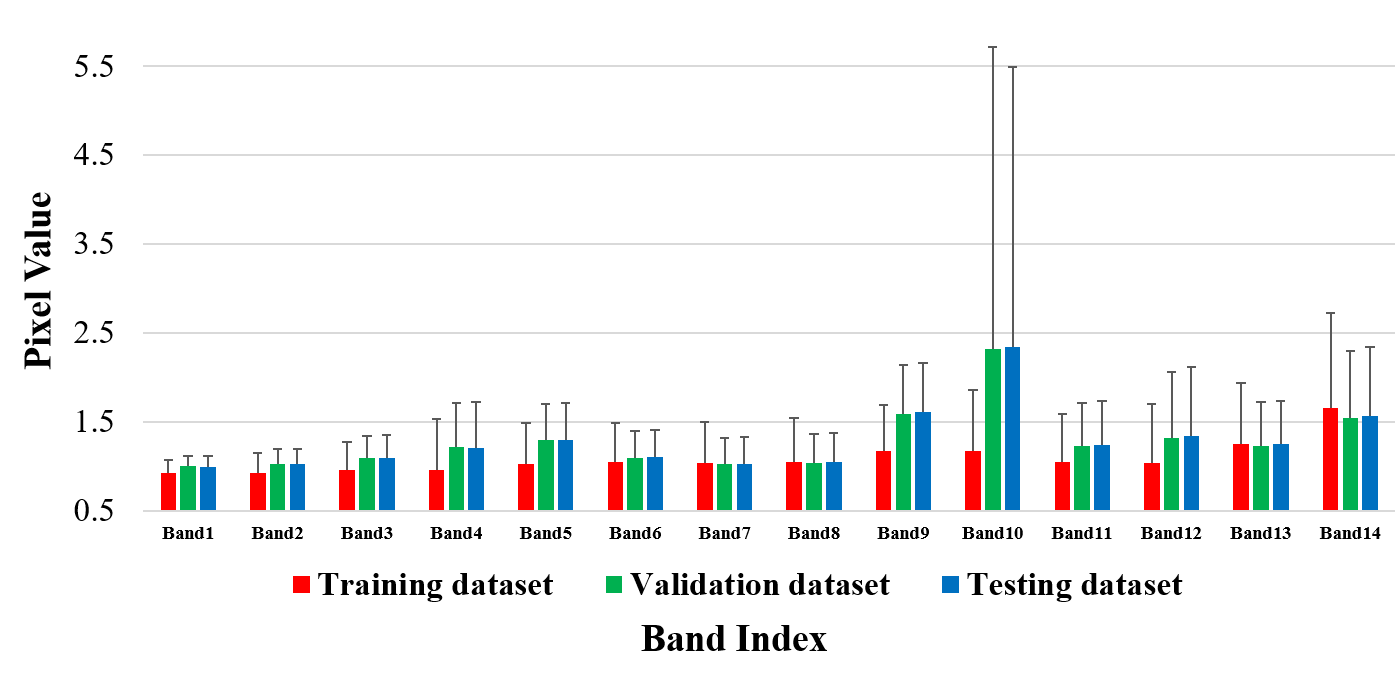}
\caption{The problem of distribution inconsistency.
The statistical results are calculated band by band, and are significantly different among the training data set, the validation data set, and the testing data set.}
\label{fig:second_problem}
\end{figure}

Distribution inconsistency is another difficult challenge for large-scale landslide detection from remote sensing imagery.
In real-world large-scale landslide detection applications, images of landslides to be detected come from all over the world. These images are collected at different times, which leads to different imaging conditions.
This spatio-temporal difference leads, in turn, to great differences in radiation values or pixel values of different remote sensing images, especially in the mountains~\cite{ZHAO2022328}, and is characterized by statistical inconsistency.
In Fig.~\ref{fig:second_problem}, the mean and standard deviation of the remote sensing images of the training, validation, and testing datasets are calculated and displayed band by band. The histogram is the mean and the error bar represents the standard deviations.
The statistical results of the training data set and the validation data set (testing data set) are quite different, causing the poor generalization performance of the model trained with the training data set data on the validation data set (testing data set).

\begin{figure*}
\centering
\includegraphics[width=\textwidth]{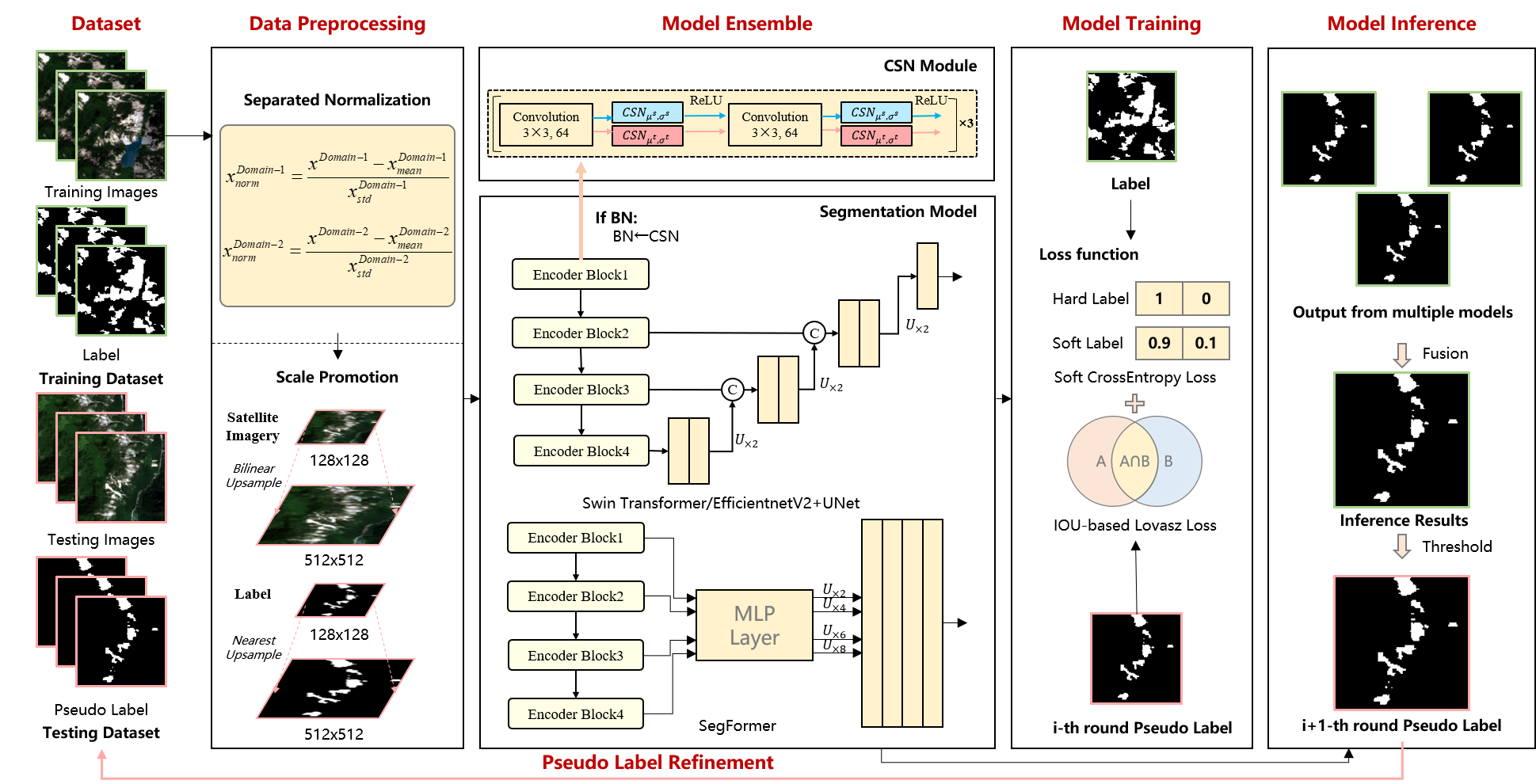}
\caption{Progressive label refinement-based distribution adaptation landslide detection framework.}
\label{fig:framework}
\end{figure*}

\subsection{Progressive Label Refinement-based Distribution Adaptation Framework}
To address the challenges of large-scale landslide detection, a progressive label refinement-based distribution adaptation framework is proposed by the first-place team for landslide detection.
As shown in Fig.~\ref{fig:framework}, the proposed framework includes data preprocessing, model ensemble, model training, model inference, and pseudo-label refinement.

\subsubsection{Data preprocessing}
Scale promotion is used to resist the weak representation caused by small landslide branches; the original images are scaled up from $128\times128$ pixels to $512\times512$ pixels.
Random flip, random rotation, and color perturbation also are adopted for data augmentation.
Color perturbation is only used for multi-spectral data, not DEM and slope data.

Separated normalization is proposed to alleviate the distribution inconsistency challenge in the data preprocessing stage, which uses the mean and variance from different domains to normalize the data.
For example, two different domains are the training domain from the training data set and the validation domain from the validation data set in the model validation stage.
The mean and standard deviation are calculated from the two datasets respectively, and then the data in the two domains are normalized respectively.
Separated normalization is similar to the normalization for cross-sensor transfer learning~\cite{WANG2022113058}, but the operation of domain-specific statistics is performed in the data preprocessing stage.

\subsubsection{Model ensemble and training}
In the segmentation model, three models are used to integrate the final landslide detection results.
The U-Decoder architecture considering multiple scales is selected as the decoder to further alleviate the small object problem, and Swin Transformer~\cite{swin} and EfficientNetV2~\cite{efv2} are selected as encoders to capture complex features of the landslide.
This framework also uses SegFormer~\cite{segformer}, which utilizes self-attention operations to fit the variant shapes of landslides, and the MLP, which is used to enhance the difficult sample features.
To further increase the generalization of the model, the batch normalization in the three segmentation models is replaced by cross-sensor normalization~\cite{WANG2022113058} to encode the statistical consistency between the training data set and the validation (testing) data set.

As for model training, Lovasz loss~\cite{loves} and an online hard example mining strategy are used to address the problem of class imbalance, and soft cross-entropy 
loss~\cite{soft} is used to solve the problem of noisy labels in the pseudo labels.

\subsubsection{Model inference and pseudo label refinement}
The probability values output by the above three models are averaged as the final prediction results in the inference stage.

To further alleviate the distribution inconsistency problem, the validation (testing) data set is used in the training process, and progressive pseudo-label refinement is proposed to generate pseudo labels for validation or testing images.
Based on the prediction of the $i$th round, pseudo labels of the $(i+1)$th round can be generated using a probability threshold of 0.7.
The models of the $(i+1)$th round can be trained by training data set and validation (testing) images with pseudo labels.
The domain adaptive consistency training and the generation of pseudo labels are performed iteratively, and the pseudo labels are refined progressively.

\subsection{Experimental Results}
We conducted a series of experiments to evaluate our proposed method on the Landslide4Sense data set.
All bands in the multi-source images were used as inputs during training and testing.
%The batch size was 16 and models were trained for 20k steps in each refinement round.

\begin{figure}[!htb]
\centering
\includegraphics[width=1.0\linewidth]{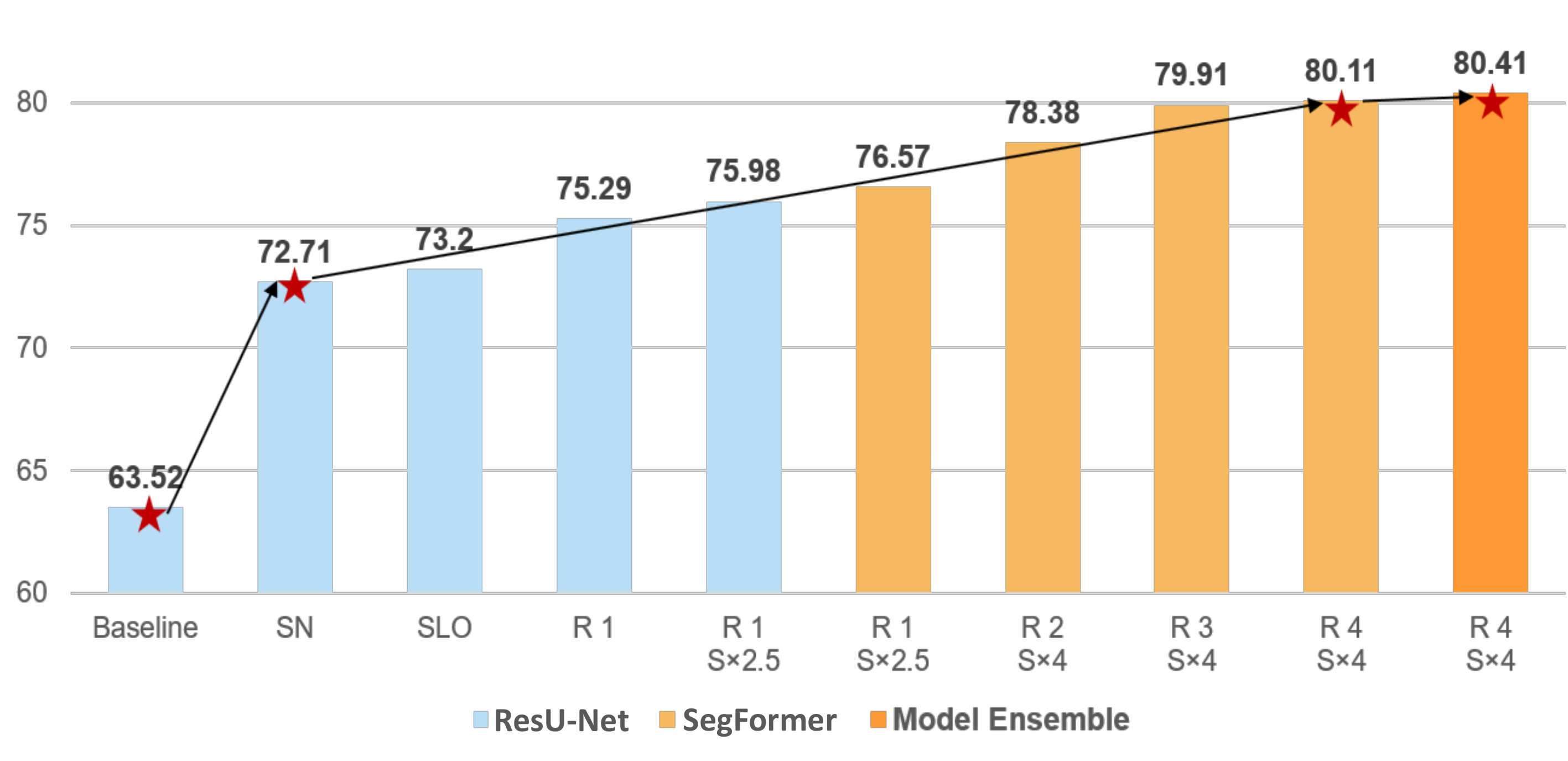}
\caption{The experimental results on the validation leaderboard. Each proposed module improves overall performance in the different aspects and is compatible with the other modules. SN -- Separate Normalization, SLO -- Soft cross-entropy loss+Lovasz loss+OHEM, S$\times$ -- Scale promotion, R -- refinement round.}
\label{fig:val_leaderboard}
\end{figure}

\begin{figure}[!htb]
\centering
\includegraphics[width=1.0\linewidth]{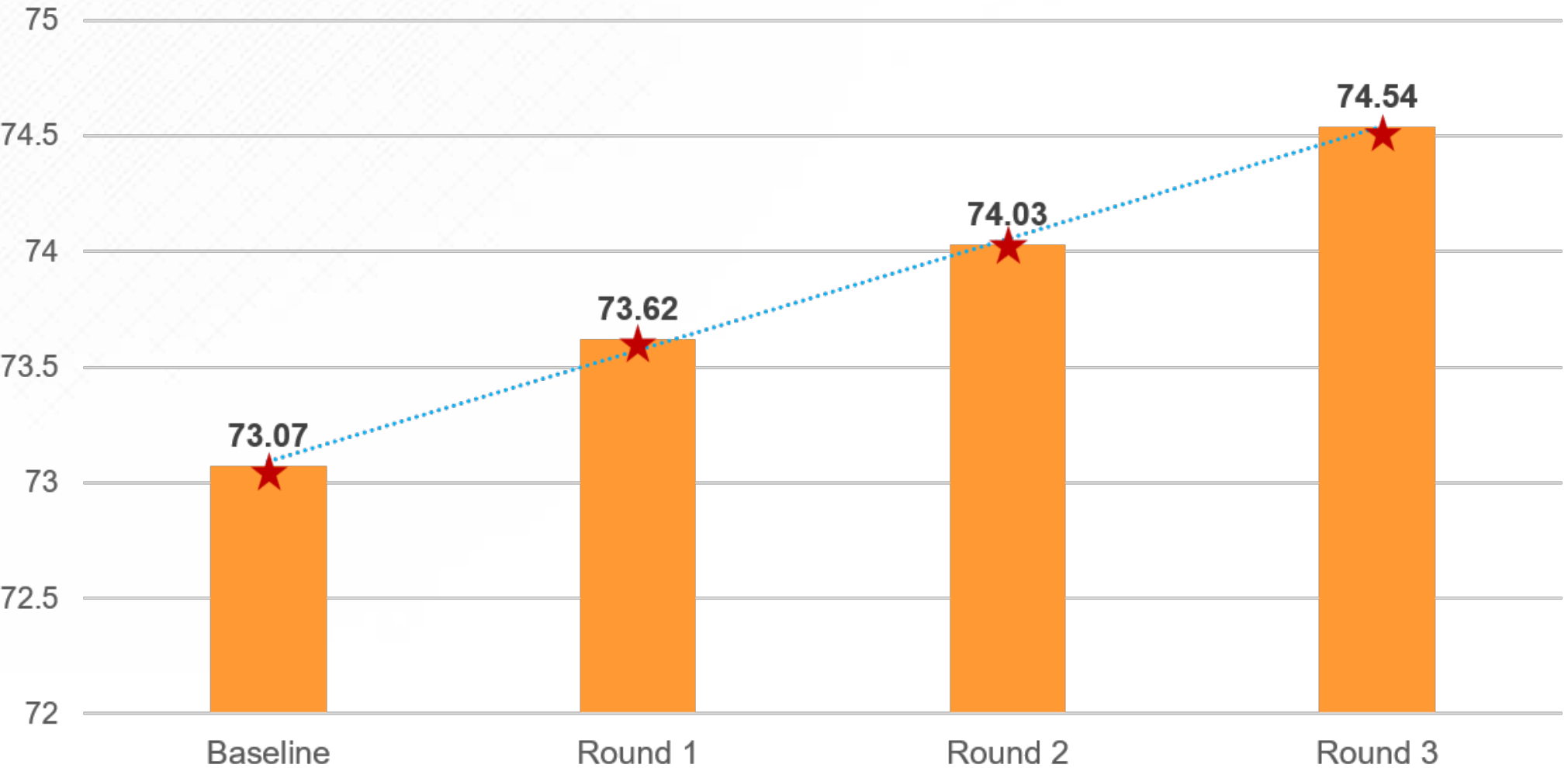}
\caption{The experimental results on the test leaderboard.}
\label{fig:test_leaderboard}
\end{figure}

The results in Fig.~\ref{fig:val_leaderboard} show that each proposed module improves landslide detection accuracy in the different aspects. In particular, the separate normalization achieves the greatest improvement, addressing the distribution inconsistencies in multi-source data.
As the number of label refinements increase, overall performance improves.
After the final round, we combine these advanced models into an ensemble to obtain the highest F1 score of 80.41\%.

As for the test leaderboard, the best model in validation experiments is utilized as the baseline and achieves an F1 score of 73.07\% (Fig.~\ref{fig:test_leaderboard}).
Consistent with the validation phase,
detection accuracy is progressively improved as the number of rounds increase. Finally, the best model obtains the highest F1 score of 74.54\%.

\section{Second-Place Team}\label{sec:seek}

\begin{figure*}[ht]
    \centering
    \includegraphics[width=\linewidth]{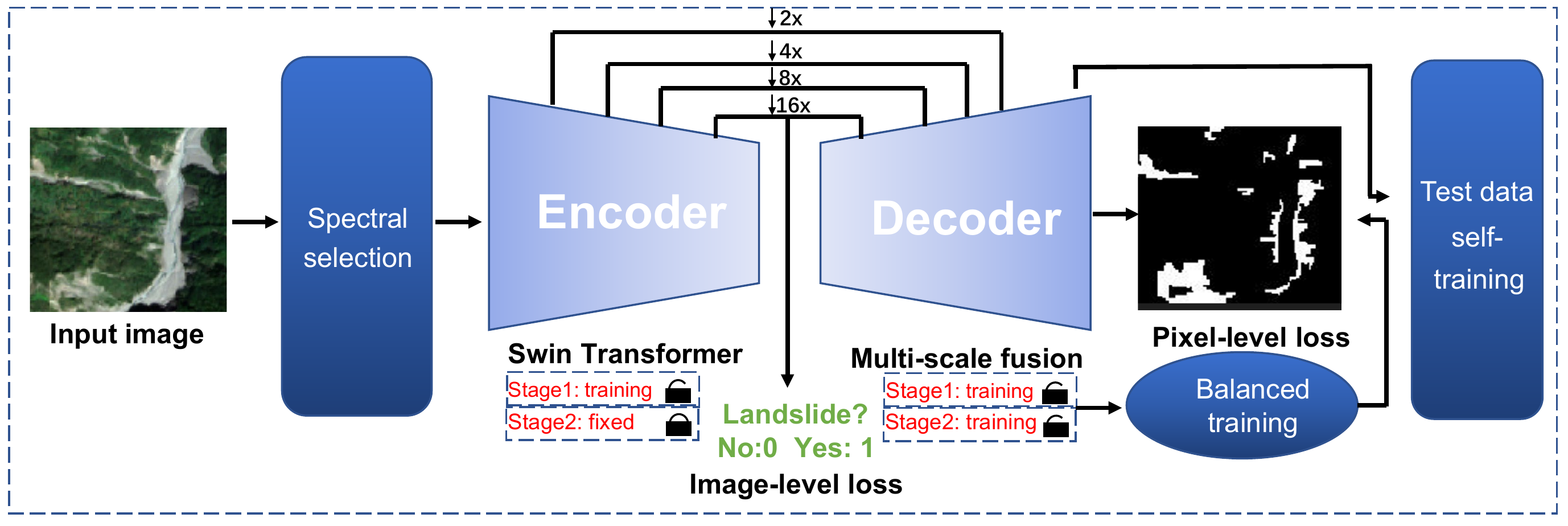}
    \caption{Model structure for landslide detection proposed by Seek team.}
    \label{seek-pipeline}
\end{figure*}

The network structure we propose for the Landslide4Sense competition is shown in Fig. \ref{seek-pipeline}. The details of each component of this proposed structure are discussed in detail below.

\subsection{Framework Introduction}
The main framework of our model is the encoder-decoder network, which uses an U-Net-like \cite{ronneberger2015u} skip connection structure and can better integrate shallow and deep features.
Influenced by the rapid development of transformer-based models in the field of computer vision\cite{swin, segformer}, we introduce Swin Transformer \cite{swin} as the encoder part in this structure.
To enable the Swin Transformer to reasonably capture the associations between landslide regions on multi-spectral data, we performed spectral selection experiments to use the spectra suitable for the self-attention mechanism.
Subsequently, in order to alleviate the imbalance problem of positive and negative samples in landslide detection, we design an unbalanced training strategy that utilizes the unbalanced loss to first train compact feature representations, and then use the feature representations to fine-tune the classifier.
Finally, we adopt a self-training strategy to further enhance the generalization of the model in the test domain.

\subsubsection{Spectral selection}
\begin{figure}
    \centering
    \includegraphics[width=\linewidth]{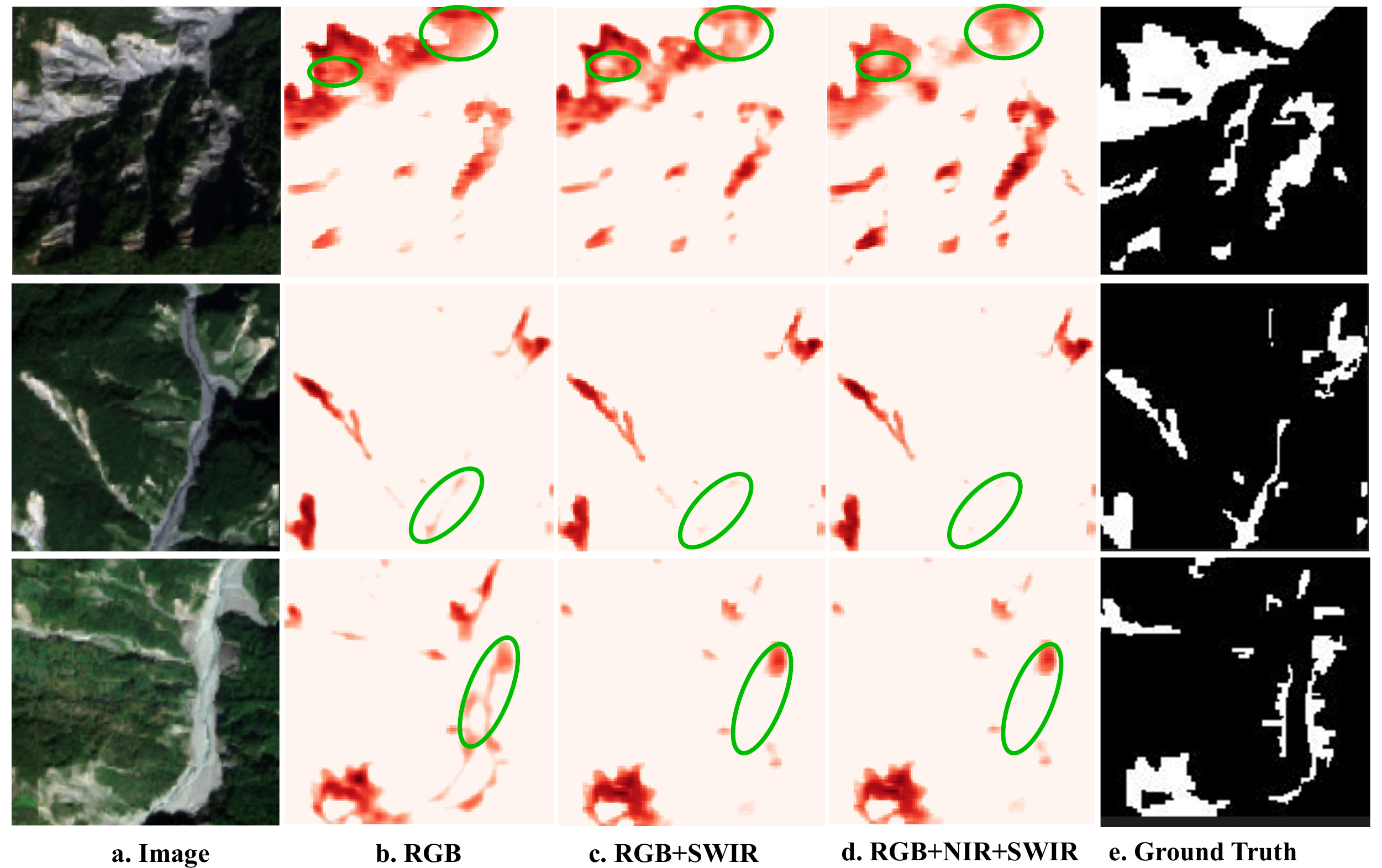}
    \caption{Visualization of the feature activation map of the Swin Transformer when inputting different spectral bands.}
    \label{seek-att}
\end{figure}

The Vision Transformer-based model performs feature aggregations using the self-attention mechanism to capture relations among pixels \cite{swin}.
 If irrelevant spectral information occupies dominant information, it will degrade the performance of the model.
 However, in the multi-spectral data, the responses of different spectra to the landslide area are quite different, and some spectra are even insensitive to landslides. Therefore, these ``unspecific'' spectra interfere with the execution of the self-attention mechanism. We performed spectral selection experiments, as shown in Table \ref{seek-bc} and find that the fully convolutional model U-Net performs better with more spectral inputs, while the Transformer model works better when only the RGB spectrum is input.
 We further visualized the negative effect on self-attention when a spectrum insensitive to landslide responses was fed into the model, as shown in Fig. \ref{seek-att}.
 Finally, we use the RGB spectra as the input to the model.

\begin{table}
  \centering
  \renewcommand{\arraystretch}{1.05}
	\caption{Spectral Selection Experiments}
  \setlength{\tabcolsep}{2.9pt}{
    \begin{tabular}{cccccc}
    \toprule
     &  & \multicolumn{3}{c}{F1 Score (\%)} \\
    Input spectral bands & Input bands  & Swin Transformer & Deeplabv3   & U-Net \\
    \hline
     RGB  & 3 &  \textbf{65.6}  &  58.0  & 59.2 \\
	  SWIR  & 3 &  55.6  &  50.2  & 52.1 \\
	  NGB  &3 &  60.8  &   \textbf{59.2}  & 58.9 \\
	  PCA \cite{martinez2001pca} &3&  49.5  &  46.8  & 52.4 \\
	  RGB + NIR  & 4 &  63.3  &  57.2  & 59.4 \\
	  RGB + SWIR  & 6 &  58.2  &  55.9  & 59.8 \\
    RGB + NIR + SWIR  &7 &  54.8  & 57.5 & 60.0 \\
    All bands  &14 &  55.8  & 57.8 & \textbf{61.1} \\
    \bottomrule
    \end{tabular}
\\
\begin{flushleft}
\scriptsize Note: In this table, the RGB denotes the red, green, and blue spectra. SWIR denotes the 3-band far infrared in Sentinel-2. NGB denotes the near-infrared, green, and blue spectra. NIR denotes the near-infrared spectral. PCA refers to the techniques \cite{martinez2001pca} of dimensionality reduction for compressing the original 14 bands into 3 bands.
\end{flushleft} 
}
\setlength{\abovecaptionskip}{0.05 cm}
  \label{seek-bc}%
\end{table}%

\subsubsection{Balanced training}
We design a two-stage training method to reduce the impact of the imbalance in the proportion of positive and negative samples.
In the first stage, both the encoder and the decoder are trained simultaneously. 
For any input samples $x_i \in R^{w \times h \times 3}$, we use weighted cross-entropy loss $\mathcal{L}_{wce}$ and Lovasz loss $\mathcal{L}_{lov}$ \cite{loves} for balanced training as follows:
\begin{equation}
\mathop {\arg \min}_{E, D} {\mathcal{L}_{wce} + \mathcal{L}_{lov} + \mathcal{L}_{ice}.}
\end{equation}
The $\mathcal{L}_{ice}$ loss is the image-level loss performed in high-level semantic features in the encoder to assist training, which is defined as follows:
\begin{equation}
 \mathcal{L}_{ice} = - \frac{1}{ |\mathcal X| } \sum_{x_{i} \in \mathcal X} \delta (y_{i}) \log MP(E(x_{i})),
\end{equation}
where $\delta$ is a pointer function. 
If there is a positive sample (landslide) in $y$, the value of $\delta$ is 1; otherwise, its value is 0. $MP(\cdot)$ is a fully connected layer with a global pooling operation. $\mathcal X$ denotes the total data set.
Optimizing the $\mathcal{L}_{ice}$ loss can increase the model's attention to landslides, since the task of finding a landslide in an image is much easier than finding where the landslide is. In order to reweight the learning of negative and positive samples, the $\mathcal{L}_{wce}$ loss is defined as follows:
\begin{equation}
 \mathcal{L}_{wce} = - \frac{1}{ |\mathcal X| } \sum_{x_{i} \in \mathcal X} \frac{N_{neg}}{N_{pos}} y_{i} \log D (E(x_{i})),
\end{equation}                                  
where $N_{neg}$ denotes the number of negative samples (non-landslides) and $N_{pos}$ denotes the number of positive samples (landslides) in any input image $x$. 

As mentioned in \cite{zhou2020bbn}, this re-weighting loss $\mathcal{L}_{wce}$ plays a positive role in balancing the feature distribution of positive and negative samples.
However, the classifier will still be biased.
Therefore, in the second stage, we fix the trained encoder $E$ and use the standard cross-entropy loss $\mathcal{L}_{ce}$ to train the decoder $D$:
\begin{equation}
\mathop {\arg \min}_{D} {\mathcal{L}_{ce} + \mathcal{L}_{ice}.}
\end{equation}
Once we have balanced feature representations, they can be further exploited to de-bias the classifier.

\subsubsection{Test data self-training}
Remote sensing imaging often faces the problem of data distribution shifts due to differences in geography and sampling time.
To fully adapt the model to the distribution of the test data, we adopt a self-training strategy \cite{zou2018unsupervised} for enhancing the generalization of the model.
We sort the output probabilities predicted in the previous stage, select the top $\lambda \%$ high-confidence pixel-level pseudo-labels, and add them to the training data for self-training.

\subsection{Experimental Results}
In this subsection, we report the performance of the balanced training and self-training methods.
Table \ref{seek-bt} shows that the two-stage balanced training method better attenuates the influence of the imbalance problem than focal loss \cite{lin2017focal} and other common methods\cite{loves}.
Table \ref{seek-st} shows that the proposed self-training method can enhance the performance of the model, and can also balance precision and recall by adjusting the value of $\lambda$.

\begin{table}
  \centering
  \renewcommand{\arraystretch}{1.05}
	\caption{Balance Training Experiments}
  \setlength{\tabcolsep}{6pt}{
    \begin{tabular}{ccc}\toprule
     &  \multicolumn{2}{c}{F1 Score (\%)} \\
    Training  & Swin Transformer & U-Net  \\
    \hline
     Normal training  & 69.8 & 63.7 \\
     Weighted cross-entropy   & 70.8 & 64.9 \\
     Focal loss \cite{lin2017focal}  & 68.2 &  61.8 \\
     Lovasz loss \cite{loves}   & 72.3 & 66.4 \\
     Balance training  & \textbf{73.9} &  \textbf{67.7} \\
\bottomrule
    \end{tabular}
	}
\setlength{\abovecaptionskip}{0.05 cm}
  \label{seek-bt}%
\end{table}%

\begin{table}
  \centering
  \renewcommand{\arraystretch}{1.05}
	\caption{Self-training Experiments with Different $\lambda$ Values}
  \setlength{\tabcolsep}{6.5pt}{
    \begin{tabular}{cccc}
    \toprule
    $\lambda$  & Precision ($\%$) & Recall ($\%$) & F1 Score ($\%$)  \\
    \hline
     - (Before ST) & 73.4 & 74.7 & 73.9 \\
     $50\%$ & 65.2 & \textbf{80.5} & 72.7 \\
     $70\%$ & 69.3 & 79.5 & 73.7 \\
     $90\%$ & 72.4 & 77.1 & 74.9\\
     $100\%$ &  \textbf{78.2} & 74.2 & \textbf{76.1} \\
\bottomrule
    \end{tabular}
	}
\setlength{\abovecaptionskip}{0.05 cm}
  \label{seek-st}%
\begin{flushleft}
\scriptsize Note: ST denotes self-training.
\end{flushleft} 
\end{table}%

In summary, the transformer-based solution we use can effectively detect landslide areas in multi-spectral remote sensing scenes.
In the future, our team argues that adaptive spectral selection or fusion technology is a necessary way to explore the performance of this transformer model further, and will become a follow-up research focus of our team.
\section{Third-Place Team}\label{sec:tanmlh}

\begin{figure*}[th]
\centering
\includegraphics[width=0.85\linewidth]{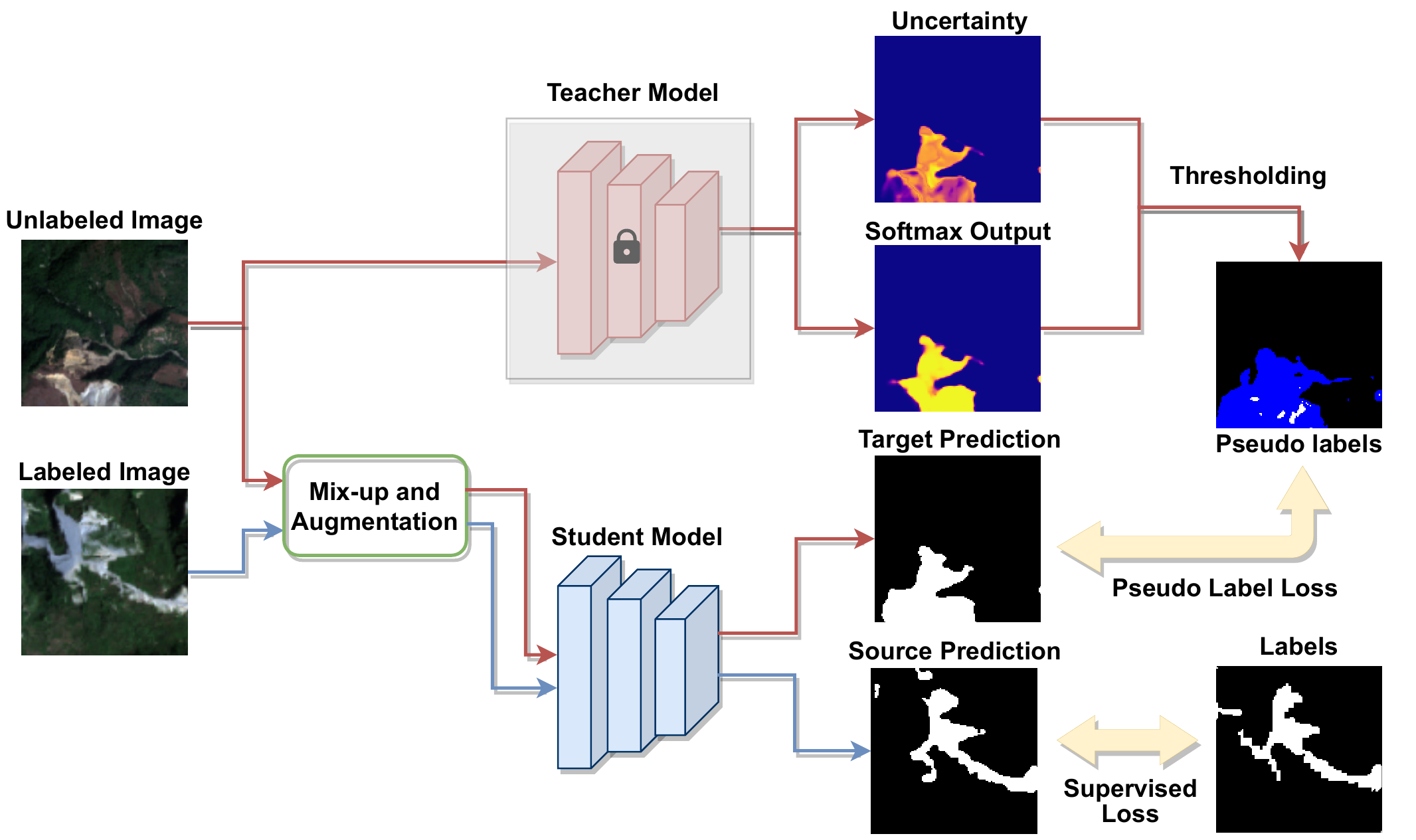}
\caption{Network architecture of the landslide detection method proposed by \emph{Tanmlh} team. The overall architecture follows a self-training scheme, which consists of a teacher model branch and a student model branch.
For the teacher model branch,
a teacher model pre-trained on the training data is applied to generate pseudo labels based on the unlabeled images, which will later be used to supervise the training of the student model.
For the student model branch,
both the labeled and the unlabeled images are input to the student model after some data augmentation and mix-up operations.
The training losses are then calculated based on both the training labels and the pseudo labels.
During the training phase, the teacher model is fixed.}
\label{fig:tanmlh_pipeline}
\end{figure*}

The solution of the third-place team is illustrated in Fig. \ref{fig:tanmlh_pipeline}.
The methodology is detailed in the following sections.

\subsection{Problem Formulation}
Technically, the landslide detection problem can be formulated as a binary semantic segmentation problem.
The training, validation, and test data set can be denoted by
$\mathcal{D}_{train}=\{x_{tr}, y_{tr}\}$,  $\mathcal{D}_{val}=\{x_{val}\}$, and $\mathcal{D}_{test}=\{x_{te}\}$,
where $x_{tr}$, $y_{tr}$, $x_{val}$, and $x_{te} \in \mathbb{R}^{H \times W}$ correspond to the training patch, training label, validation patch, and test patch, respectively. Here $H$ and $W$ denote the data sets' spatial size.
The goal of the landslide detection task is to train a semantic segmentation model on $D_{train}$ and $D_{val}$, so that the best performance can be achieved on $\mathcal{D}_{test}$.
Since the data are collected from different regions across the world,
improving the exploitation of the unlabeled validation data can be beneficial to mitigate the domain gap between all the labeled and unlabeled data.
To this end, we propose to incorporate a mixed supervised loss $\mathcal{L}_{sup}^{mix}$ and a mixed pseudo-label loss $\mathcal{L}_{pse}^{mix}$ to train the network:
\begin{equation}
    \mathcal{L} = \mathcal{L}_{sup}^{mix} + \mathcal{L}_{pse}^{mix}.
\end{equation}
The detailed formulation of $\mathcal{L}_{sup}^{mix}$ and $\mathcal{L}_{pse}^{mix}$ will be given in Section \ref{sec:mix_up}.

\subsection{Supervised Losses}
A combination of the cross-entropy loss $\mathcal{L}_{cet}$ and the Jaccard loss $\mathcal{L}_{jac}$ \cite{polak2009evaluation} are used as the supervised losses:
\begin{equation}
    \mathcal{L}_{sup} = \mathcal{L}_{cet}(\mathcal{M}_{s}(x_{tr}), y_{tr})
    + \mathcal{L}_{jac}(\mathcal{M}_{s}(x_{tr}), y_{tr}).
\end{equation}
Here $\mathcal{M}_{s}(\cdot)$ denotes the mapping function defined by the student model $\mathcal{M}_{s}$.

\subsection{Self-training}
\label{sec:st}
The authors propose a self-training strategy \cite{zou2018unsupervised} to exploit the unlabeled data.
First, the teacher model $\mathcal{M}_{t}$ will be trained solely on the training data $\mathcal{D}_{train}$. Then it will be used to generate pseudo labels on the unlabeled data $\mathcal{D}_{val}$ to supervise the student model $\mathcal{M}_{S}$.

However, the raw predictions from $\mathcal{M}_{t}$ are likely to be incorrect.
To prevent the student model from overfitting to those wrong predictions,
a pseudo-label selection strategy is needed to filter out misclassified pixels.

To achieve this, 
the Monte Carlo dropout strategy \cite{gal2016dropout} is first used to generate an uncertainty map for each unlabeled image patch.
%\begin{comment}
More specifically, the unlabeled validation patch $x_{val}$ is input to the teacher model $\mathcal{M}_{t}$ for $10$ different runs.
During each run, 
a dropout layer with $0.3$ dropping rate is applied after the first convolution layer to disturb the network.
The variances of $10$ different outputs are then calculated as the uncertainty map.

%\end{comment}
Next, the uncertainty map is used to mask out those uncertain predictions from the teacher model $\mathcal{M}_{t}$.
Inspired by class-balanced self-training (CBST) \cite{zou2018unsupervised},
the selection process is conducted in a class-wise manner,
which means the top $90\%$ of the background pixels and top $70\%$ of the landslide pixels with the lowest uncertainty will be selected as the pseudo labels.
Meanwhile, the other predictions with higher uncertainty will be ignored when calculating the 
 losses.

To this end, the pseudo label loss $\mathcal{L}_{pse}$ can be formulated by:
\begin{equation}
    \mathcal{L}_{pse} = \mathcal{L}_{cet}(\mathcal{M}_{s}(x_{te}), \hat{y}_{te}) + \mathcal{L}_{jac}(\mathcal{M}_{s}(x_{te}), \hat{y}_{te}).
\end{equation}
Here $\hat{y}_{te}$ corresponds to the pseudo labels of $x_{te}$ generated by the teacher model and followed by the pseudo label selection process.

\subsection{Mix-up Strategy}
\label{sec:mix_up}
To prevent overfitting and further improve the generalizability of the landslide detection model,
a mix-up strategy \cite{zhang2017mixup} is applied to both the labeled and the unlabeled data.
Given a batch of the training data $x_{tr}$ and the validation data $x_{val}$,
the mixed data can be achieved by their linear mixing:
\begin{equation}
\begin{aligned}
    \tilde{x}_{tr} = \lambda x_{tr}^{i} + (1 - \lambda) x^{j}_{tr}, \\
    \tilde{x}_{val} = \lambda x_{val}^{i} + (1 - \lambda) x^{j}_{val}.
\end{aligned}
\end{equation}
Here $x^{i}$ and $x^{j}$ are two image patches from the corresponding data set, and $\lambda$ is the mixing coefficient randomly sampled from a beta distribution during each training step.
After applying the mix-up strategy,
the supervised and pseudo label losses can be reformulated as
\begin{equation}
\begin{aligned}
    \mathcal{L}_{sup}^{mix} &= \lambda \mathcal{L}_{sup}(\tilde{x}_{tr}, y_{tr}^{i}) + (1- \lambda) \mathcal{L}_{sup}(\tilde{x}_{tr}, y_{tr}^{j}), \\
    \mathcal{L}_{pse}^{mix} &= \lambda \mathcal{L}_{pse}(\tilde{x}_{te}, \hat{y}_{te}^{i}) + (1- \lambda) \mathcal{L}_{pse}(\tilde{x}_{te}, \hat{y}_{te}^{j}).
\end{aligned}
\end{equation}
By training on mixed images,
the model will be less likely to be overconfident about its predictions,
and hence better generalize to the unseen data.

\subsection{Post-processing}
The dense conditional random field (DenseCRF) \cite{krahenbuhl2011efficient}
technique is applied to the model's output as post-processing.
This step helps to better match the predicted landslide contours with the ground truths. Finally, the best model obtains the highest F1 score of 73.5\%.

\section{Special Prize Team}\label{sec:sklgp}

\begin{figure*}[ht]
    \centering
    \includegraphics[width=\linewidth]{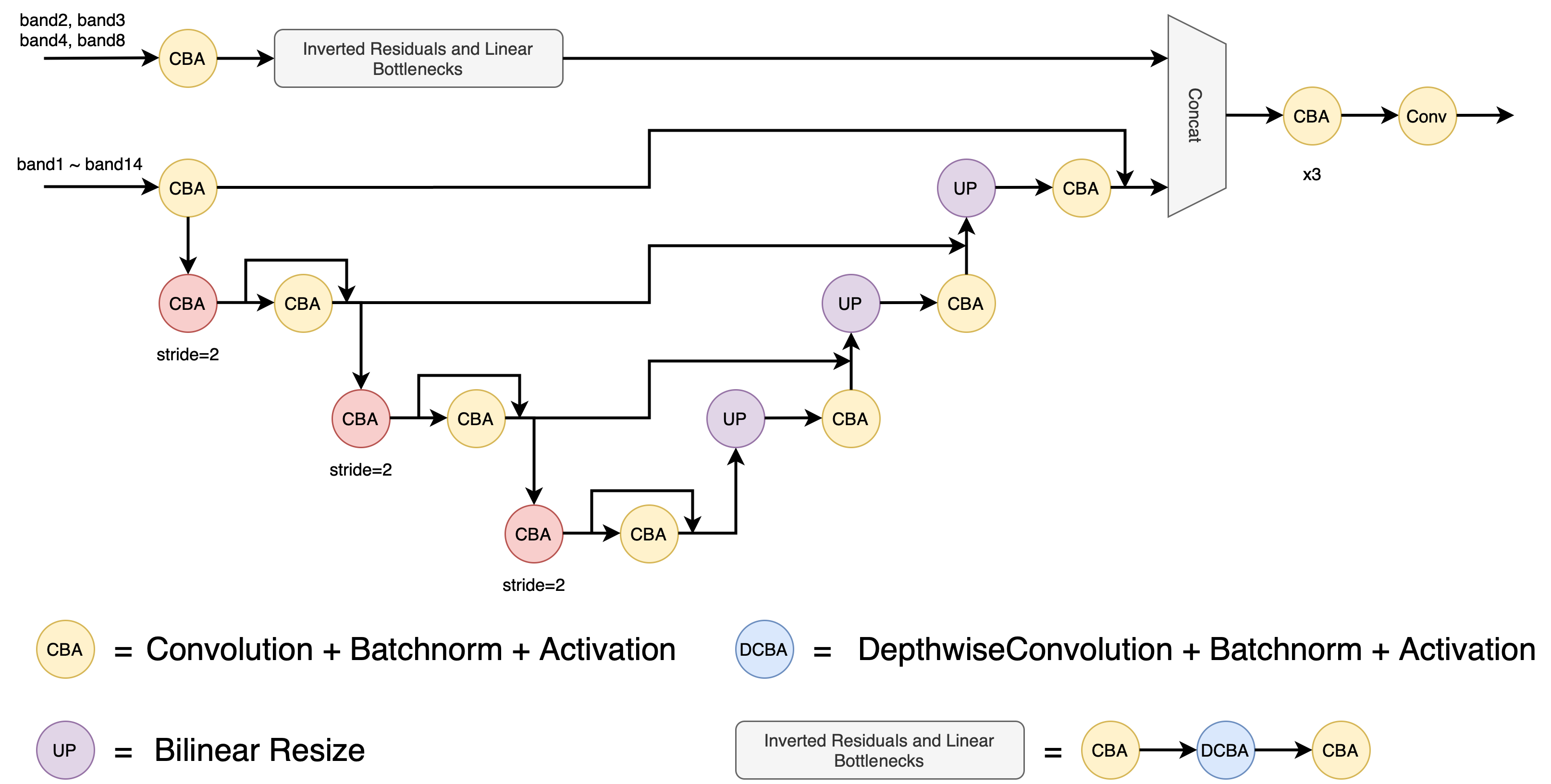}
    \caption{The structure of the Multi-spectral U-Net. The inputs of the upper branch consist of band2, band3, band4, and band8, which have 10 meters resolution, and the inputs of the lower branch mix all the 14 bands, including 10, 20, and 60 meters resolutions. Specifically, we implement the downsampling layer by using the convolution with stride two and using the bilinear resize for the upsampling layer, and all the skip connection operations are additive. CBA means the sequential block of convolution, batch norm, and activation.}
    \label{multispectral-model}
\end{figure*}

Landslide4Sense provides data with 14 bands, while most deep learning semantic segmentation models, such as 
\cite{FCN,ronneberger2015u,Chen2017RethinkingAC,chen2018encoder}, require 
an RGB image as the input. This means we cannot utilize pre-trained weights to improve the model performance 
and shorten training time. On the Landslide4Sense data set, we try three types of models, U-Net, Deeplabv3, and Deeplabv3+, 
but none of these models yields a very high performance, with F1 scores of only 65\%, 66\%, and 67\%, respectively. 
So, we explore the use of multi-spectral satellite imagery for the deep learning-based landslide segmentation task.

\subsection{Multi-spectral U-Net}
Considering the different resolution bands of the imagery in the Landslide4Sense dataset, we introduce a novel model 
called Multi-spectral U-Net, which has two input branches for the different resolution inputs. The model structure is 
illustrated in Fig. \ref{multispectral-model}. Multi-spectral U-Net comprises two branches, the High Resolution Branch 
(upper part) and the General Resolution Branch (lower part), whose features will be merged, and then contribute jointly to the final segmentation prediction. 

The High Resolution Branch was used for the data with high resolution, which can yield refined feature maps 
containing more marginal information. Specifically, we implement this branch by using the Inverted Residuals and 
Linear Bottlenecks introduced in the MobileNetV2 \cite{Sandler2018MobileNetV2}, and consisting of two point-wise convolution layers 
and one depth-wise convolution layer. To avoid a dramatic increase in the dimensions of the feature maps, from 4 to 128 dimensions, we first apply two 
simple convolution layers. The feature dimensions will expand then 
recovered to the original dimension after the depth-wise convolution layer. Additionally, there is no downsampling layer 
in the branch, as the only aim of this branch is to extract additional marginal information in order to get a better 
segmentation prediction.

In the General Resolution Branch, we apply some modifications to the original U-Net \cite{ronneberger2015u}. U-Net is an expandable segmentation 
model that has a symmetrical architecture; this kind of architecture has been widely used for other segmentation 
tasks. It is very convenient to replace some implementations of the U-Net, which is the main reason we chose it for our model. The specific modifications are as follows. First, we reduce the number of the 
downsampling layers due to the limited size ($128 \times 128$ pixels) of the input image. To ensure the smallest feature size is at least 
$16 \times 16$, we use only three downsampling operations in the U-Net model. Secondly, 
skip-connection introduced in the ResNet is widely used in the model to mitigate the 
vanishing gradient problem. Finally, we update the activation function to SMU \cite{Biswas2021SMUSA}, which can improve model performance 
without performance loss on inference speed, as shown in Eqs.  \eqref{erf}, and \eqref{smu}.

\begin{equation}\label{erf}
    erf(x) = \frac{2}{\sqrt{\pi}} \int_{0}^x{e^{-t^2}dt}.
\end{equation}

\begin{equation}\label{smu}
    SMU(x) = \frac{x(1 + \alpha) + x(1 - \alpha)erf(\mu(1-\alpha)x)}{2}.
\end{equation}

In the Multi-spectral U-Net model, we input all 14 bands to the General Resolution Branch and only 10 meters 
resolution bands (band2, band3, band4, and band8) to the High Resolution Branch. In order to balance the feature dimensions of two 
branches, we make the High Resolution Branch and the General Resolution Branch have the same output shape, 
$128 \times 128 \times 128$. The features from two branches will be concatenated to a feature map in the shape of $128 \times 128 \times 256$, 
which is used for the final pixel-level prediction.

\subsection{Experiments}

\begin{figure}
    \centering
    \includegraphics[width=0.85\linewidth]{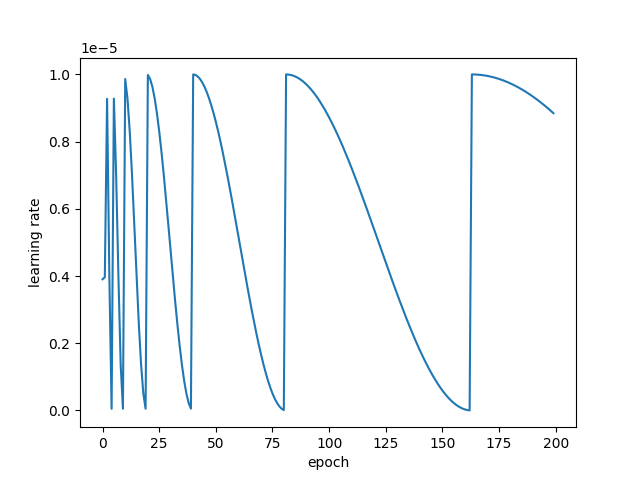}
    \caption{The cosine learning rate with warmup and restart schedule. 
    In detail the parameters are: warmup epochs 2, restart multiplier 2, init learning rate 0.00001, and minimum learning rate 0.}
    \label{learningrate}
\end{figure}

We trained the model with NVIDIA GeForce RTX 3090 GPU and Intel(R) Core(TM) i7-7800X CPU @3.50GHz. To compare 
the performance of the three models more clearly, we use a batch size of 8, 
the Adam optimizer, warmup, and restarted cosine learning rate (shown in Fig. \ref{learningrate}) and the cross-entropy loss. 
% Significantly, we try to use some augmentation strategies, such as random 
% flip and rotate, but these strategies do not have improvement in the performance of the model, so we train the model without 
% any augmentation.

\begin{figure}
    \centering
    \subfloat{
    \includegraphics[width=1.8cm]{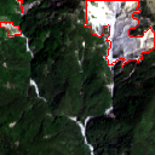}
	}
    \hspace{0.8mm}
    \subfloat{
    \includegraphics[width=1.8cm]{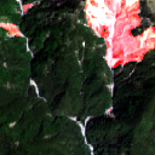}
	}
    \hspace{0.8mm}
    \subfloat{
    \includegraphics[width=1.8cm]{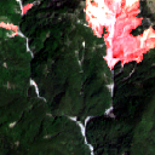}
	}
    \hspace{0.8mm}
	\subfloat{
    \includegraphics[width=1.8cm]{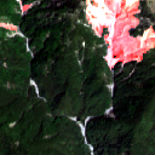}
	}

    \vspace{-2.5mm}
    \subfloat{
    \includegraphics[width=1.8cm]{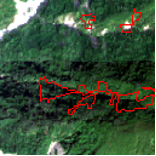}
	}
    \hspace{0.8mm}
    \subfloat{
    \includegraphics[width=1.8cm]{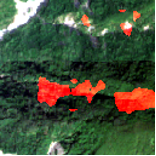}
	}
    \hspace{0.8mm}
    \subfloat{
    \includegraphics[width=1.8cm]{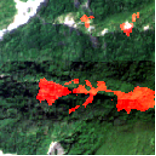}
	}
    \hspace{0.8mm}
	\subfloat{
    \includegraphics[width=1.8cm]{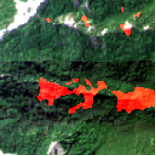}
	}

    \vspace{-2.5mm}
    \subfloat{
    \includegraphics[width=1.8cm]{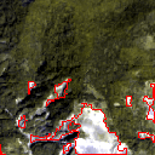}
	}
    \hspace{0.8mm}
    \subfloat{
    \includegraphics[width=1.8cm]{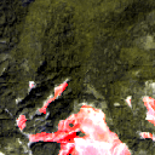}
	}
    \hspace{0.8mm}
    \subfloat{
    \includegraphics[width=1.8cm]{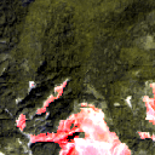}
	}
    \hspace{0.8mm}
    \subfloat{
    \includegraphics[width=1.8cm]{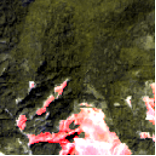}
	}

    \vspace{-2mm}
    \flushleft{
    \footnotesize{
        \hspace{0.5mm}
        \textbf{Ground Truth}
        \hspace{5mm}
        \textbf{Deeplabv3+}
        \hspace{10mm}
        \textbf{U-Net}
        \hspace{8mm}
        \textbf{Multi-spectral\\ \hfill U-Net\quad}
    }
    }
    \vspace{-2mm}

    \caption{The landslide segmentation results for some representative data in the validation dataset.}
    \label{segmentation-sample}
\end{figure}

We split the official training dataset into two parts, with 3539 images for training and 260 images for testing. 
Then, we compared the Multi-spectral U-Net performance with Deeplabv3+ and U-Net on it, after each model is trained 
with 200 epochs. In terms of recall and F1 score, Multi-spectral U-Net is significantly higher than the other two models, 
but its precision is lower than that of U-Net. Significantly, the precision of Deeplabv3+ is dramatically lower than Multi-spectral U-Net and U-Net, 
and we think the potential reason is that a large number of downsampling layers lead to the loss of marginal information.

% The recall, precision and f1 score in the table \ref{tab:evaluation} is the best results on the validation dataset 
% during the training process. 
% In terms of recall and f1 score, Multispectral U-Net achieves 75.41 and 77.83 respectively, which are significant higher then other 
% two models, but the precision is 1.41 lower than U-Net (81.82). 
% Significantly, the precision of Deeplabv3+ is only 74.22, which is dramatically lower than Multispectral U-Net and U-Net, 
% and we think the potential reason is that the large number of downsampling layers lead to the loss of marginal information.

% \begin{table}[h]
%     \centering
%     \caption{Evaluation results on the LandSlide4Sense dataset.}
%     \label{tab:evaluation}
%     \begin{tabular}{lccc}
%         \toprule              &Recall             &Precision          &F1 Score\\
%         \midrule
%         Deeplabv3+            & 72.54             & 74.22             & 73.37 \\
%         U-Net                 & 71.34             & \textbf{81.82}    & 76.22 \\
%         Multispectral U-Net   & \textbf{75.41}    & 80.41             & \textbf{77.83} \\
%         \bottomrule
%     \end{tabular}
% \end{table}

For a better understanding of the different models' performance on the validation dataset, we analyze the prediction 
segmentation results and choose three representative examples in Fig. \ref{segmentation-sample}. 
In the first example, the landslide segmentation results of the three models are similar, and it is clear that the Deeplabv3+ 
tends to predict a wide range of area but does not have very refined edge information about the landslide. In other words, this may be the reason that the Deeplabv3+ has a higher recall than U-Net, but the precision is significantly lower.
We can not directly see the landslide from the image in the second example; however, all three models can predict the landslide very well, which means all the bands besides the RGB bands (band2--band4) also contribute to the final prediction.
The third example is a very complex landslide scenario, in which we can clearly view the superior performance of the Multi-spectral U-Net.

In the test phase of the LandSlide4Sense competition, we use the well-trained model to predict the validation data set first and 
get annotations from the prediction result. Then, we get a new training data set by combining the annotated validation data set and the old 
training data set. Finally, after training the Multi-spectral U-Net on the new training data set, we get an F1 score of 71.29\% in the test set.

\section{Conclusions}\label{sec:concl}

% topic summary
In recent decades, remote sensing techniques have been predominantly used for natural hazard-related applications, i.e., landslide detection. There are many advantages to using Earth observation and remote sensing products in these applications, but the most critical one is their timeliness and objectivity. Early detection is vital for a rapid response and effective management of the consequences of a landslide event. Due to the increasing number and quality of space-borne sensors, the remote sensing community has recently had access to high-quality images with a higher spatial-temporal resolution. In light of the improved availability of data, attention has turned towards the methodologies for retrieving information and knowledge from the data itself \cite{lang2021multi}. Therefore, there has been a great desire to replace the use of experts' knowledge-based physical methodologies with automatic interpretation methods of remote sensing images.

Although promising results have been obtained by DL models for a wide range of remote sensing applications, the need for solutions to landslide detection challenges such as extracting landslides from remote sensing data has only been brought to the attention of the machine learning and computer vision communities in recent years. The solutions, however, have only been implemented at the local level and have followed a common procedure that includes training the DL model using an annotated data set of landslides covering a relatively small area \cite{Ghorbanzadeh2021,Prakash2021}. The local level is taken into consideration for several reasons related to how model generalization handle high-level issues, such as the impacts caused by different triggers, the types of mass movements, and the geology and morphology of the region, as well as the source of inventory data sets and the method in which they were developed. The landslide inventory data sets that are used for training modern DL models are usually created based on manual or knowledge-based physical semi-automated methods. Thus, implementing such methods for semantic annotations and creating inventory data sets at a large scale is generally a tedious and expensive process. In preparing a precise inventory of landslides, an even greater amount of work is required since it involves not only the analysis of one image but also a comparison of two images from the pre- and post-event for each case study area. Therefore, it is very unlikely that landslide inventory data sets with highly accurate annotations can be found on a large scale. As a result of the lack of these data sets, serious concerns about the performance of currently available landslide detection DL solutions are warranted, particularly, when applied directly to a new case study area that has not yet been investigated. To address all the above-mentioned issues, the L4S competition has been organized by the IARAI and provides a globally distributed landslide inventory data set. The competition promotes development and demonstration of innovative algorithms for automatic landslide detection using remote sensing images throughout the world, as well as providing fair and objective comparisons of different DL solutions for automatic landslide detection. 

This paper presents a summary of the top winners of the 2022 L4S competition. The competition was dedicated to developing DL solutions for solving unsolved challenges in the detection of landslides using remote sensing images collected from various regions around the world. Different strategies and algorithms were brought to light by our winning teams. The first-ranked team identified three main challenges: a large number of small landslides and the huge class imbalance between landslides and non-landslides, as well as the distribution inconsistency of the landslides in the study areas and, consequently, the image patches. In addressing these challenges, they conducted a series of experiments to obtain the competition's highest F1 score value of 73.07\%. For the weak representation of small landslides, they applied a scale promotion of original image patches from $128 \times 128$ pixels to $512 \times 512$ pixels. This team integrated three models of  Swin Transformer, EfficientNetV2, and SegFormer by emphasizing self-attention operations. Further landslide detection improvements were effected by the second-place team using the Swin Transformer as the encoder part and the self-attention mechanism. In addition, a self-training strategy was used to enhance the generalization of their proposed model on the competition's test data. To overcome the imbalance between landslide and non-landslide classes, the first-place team adopted and applied the Lovasz loss and online hard example mining strategy. An unbalanced approach to training, however, led to the second team's success. The third-place team proposed an integrated approach of a mixed supervised loss and a self-training consisting of pseudo labels and the Monte Carlo dropout strategy to train their network for landslide detection. Using DenseCRF, this team post-processed the network's outputs to improve the borders of landslides. The special prize team introduced a 
multi-spectral U-Net inspired by MobileNetV2 to handle the multi-spectral Sentinel-2 and ALOS PALSAR data for landslide detection provided by the competition. As part of the competition's test phase, they generated annotations for the validation data set using the well-trained model, and by adding new labeled data to the training data set, they trained their introduced U-Net. The DL solutions provided by the competition's four winners were presented by the corresponding authors at the CDCEO 2022 workshop as a satellite event at  IJCAI-ECAI 2022, the 31st International Joint Conference on Artificial Intelligence, and the 25th European Conference on Artificial Intelligence. 

The data remain accessible after the L4S competition and the \textit{Future Development Leaderboard} for future evaluation at \url{https://www.iarai.ac.at/landslide4sense/challenge/} is active to allow further research developments and contributions. In this way, anyone can submit landslide detection results on the test data set, make comparisons of their performance to that of other users, and, ideally, enhance the accuracy presented in this outcome paper. It is noteworthy that L4S was the first competition to be based on multi-source satellite imagery for landslide detection and had a significant impact on this field; furthermore, participants agree that the competition  is also an extremely interesting challenge from a computer vision and machine learning perspective. 

As the consequences of climate change pose an accelerating quantity and range of challenges to the world's scientists, they may not have sufficient time and resources to generate landslide inventory data sets based on fieldwork. Yet modern DL solutions, particularly those based upon such a large source of remote sensing data, must be able to cope with monitoring natural hazards and risk assessment. Therefore, developing innovative DL solutions and training them on a global data set will be crucial to generating timely information from remote sensing data for future landslide events. The L4S 2022 data provide a valuable benchmark data set for evaluating all new DL algorithms developed for landslide detection, and the algorithms developed as part of the L4S competitions will, it is hoped, inspire development of increasingly efficient and accurate algorithms.

\section*{Acknowledgments}

The authors would like to thank Alina Mihai, Miriam V\'azquez, Gabriel Fratica, Toma Furdui, Gero Nikolov, Ezgi \"Ozkan, Pedro Herruzo, Christian Eichenberger, David Kreil, Michael Kopp, Sepp Hochreiter, Cees van Westen, Martin Rutzinger, and Changlin Wang for their valuable help in the organization of Landslide4Sense 2022 competition, and the anonymous referees for their valuable comments/suggestions that have helped us improve an earlier version of the manuscript. This research was funded by the Institute of Advanced Research in Artificial Intelligence (IARAI) GmbH. 

\bibliography{L4S,kingdrone,seek,tanmlh,sklgp}

%\begin{biographynophoto}
%{XXX}
%obtained ...
%\end{biographynophoto}


\end{document}